\documentclass[10pt,twocolumn,letterpaper]{article}

\usepackage[pagenumbers]{iccv} 

\definecolor{iccvblue}{rgb}{0.21,0.49,0.74}
\usepackage[pagebackref,breaklinks,colorlinks,allcolors=iccvblue]{hyperref}
\usepackage{multirow}
\usepackage{colortbl}
\usepackage{graphicx}
\usepackage{booktabs}
\usepackage{xcolor}
\usepackage{verbatim}
\usepackage{arydshln}
\usepackage{array}
\usepackage{makecell}
\usepackage{amssymb}
\usepackage{pifont}
\newcommand{\tabincell}[2]{\begin{tabular}{@{}#1@{}}#2\end{tabular}}

\definecolor{mygray}{gray}{0.9}

\title{\emph{OmniSTVG}: Toward Spatio-Temporal Omni-Object Video Grounding}

\author{
    Jiali Yao$^{1}$\;\;\; 
    Xinran Deng$^{1}$\;\;\;
    Xin Gu$^{1}$\;\;\;
    Mengrui Dai$^{2}$\;\;\;
    Bing Fan$^{3}$\;\;\;
    Zhipeng Zhang$^{4}$ \\
    Yan Huang$^{3}$ \;\;\;
    Heng Fan$^{3\dagger}$\;\;\; 
    Libo Zhang$^{5\dagger}$ \\
    $^{1}$University of Chinese Academy of Sciences\;\;
    $^{2}$North China University of Technology \\ $^{3}$University of North Texas \;\;
    $^{4}$Shanghai Jiao Tong University  \\$^{5}$
Institute of Software Chinese Academy of Sciences
}

\begin{document}

\twocolumn[{%
\renewcommand\twocolumn[1][]{#1}%
\maketitle
\begin{center}
    \centering
    \captionsetup{type=figure}
    \includegraphics[width=0.955\linewidth]{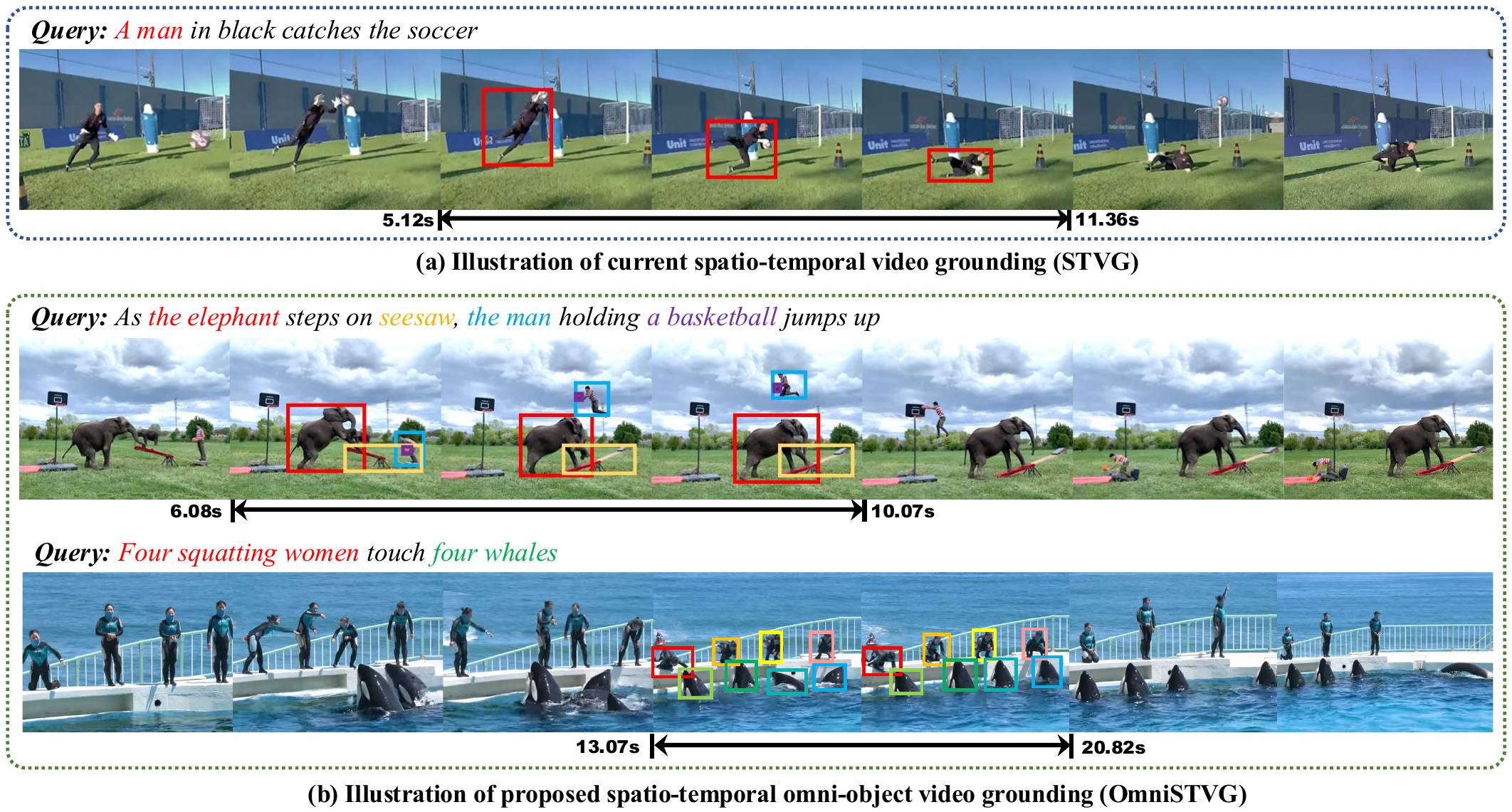}   
    \captionof{figure}{Illustration and comparison of existing \emph{STVG} that localizes a single object in the query (image (a) in the top) and our \emph{OmniSTVG} locating all objects in the query (image (b) in the bottom). The object in the textual query and its corresponding spatio-temporal tube in the video is highlighted using the same color (please notice that, the tubes for ``\emph{women}'' and ``\emph{whales}'' in (b) are displayed in different colors for better distinction). \emph{Best viewed in color and by zooming in for all figures throughout the paper}.} 
    \label{fig:task_compare}
\end{center}%
}]

\maketitle

\def\thefootnote{$^{\dagger}$}\footnotetext{Equal advising and co-last authors.}
\def\thefootnote{\arabic{footnote}}

\begin{abstract}
In this paper, we propose spatio-temporal omni-object video grounding, dubbed \textbf{OmniSTVG}, a new STVG task that aims at localizing spatially and temporally \emph{all} targets mentioned in the textual query from videos. Compared to classic STVG locating only a single target, OmniSTVG enables localization of not only an arbitrary number of text-referred targets but also their interacting counterparts in the query from the video, making it more flexible and practical in real scenarios for comprehensive understanding. In order to facilitate exploration of OmniSTVG, we introduce \textbf{BOSTVG}, a large-scale benchmark dedicated to OmniSTVG. Specifically, our BOSTVG consists of 10,018 videos with 10.2M frames and covers a wide selection of 287 classes from diverse scenarios. Each sequence in BOSTVG, paired with a free-form textual query, encompasses a varying number of targets ranging from 1 to 10. To ensure high quality, each video is manually annotated with meticulous inspection and refinement. To our best knowledge, BOSTVG is to date the first and the largest benchmark for OmniSTVG. To encourage future research, we introduce a simple yet effective approach, named \textbf{OmniTube}, which, drawing inspiration from Transformer-based STVG methods, is specially designed for OmniSTVG and demonstrates promising results. By releasing BOSTVG, we hope to go beyond classic STVG by locating every object appearing in the query for more comprehensive understanding, opening up a new direction for STVG. Our benchmark, model, and results will be released at \url{https://github.com/JellyYao3000/OmniSTVG}.
\end{abstract}

\section{Introduction}
\label{sec:intro}

Spatio-temporal video grounding (STVG)~\cite{zhang2020does} has been one of the crucial problems in multimodal video understanding. Given a free-form textual query, it aims at locating the target of interest in space and time within the video (see Fig.~\ref{fig:task_compare} (a)). Owing to its key applications, including human-machine interaction, robotics, \emph{etc}., STVG has gained increasing interest in recent years (\eg,~\cite{su2021stvgbert,gu2024context,yang2022tubedetr,jin2022embracing,lin2023collaborative,wasim2024videogrounding,gu2025knowing}).

While significant progress has been witnessed, localizing only a single object, as it is done in current STVG, is \emph{insufficient} for video understanding in many real-world scenarios. For instance, in daily life, since an event or an activity often involves various objects (see Fig.~\ref{fig:task_compare} (b)), it is common that a textual query contains \emph{multiple} targets of interest (\eg, ``\emph{elephant} and \emph{man}'' and ``\emph{four women}'' in queries in Fig.~\ref{fig:task_compare} (b)). For such queries, it is essential to localize \emph{every} queried objects, spatially and temporally, within the video. Yet, existing STVG localizes only a single target in query (see Fig.~\ref{fig:task_compare} (a)), and hence is restricted in multi-objective query localization, degrading practicability. To alleviate this, a straightforward solution is to apply the STVG model repeatedly for multiple single-object textual queries. Nonetheless, this significantly increases the computational burden, thus leading to the lack of flexibility and scalability for current STVG in practice.

In addition to the restriction in locating multiple queried targets, another limitation of current STVG is the \emph{ignorance} of interacting counterparts for queried objects. In practice, the object of interest is usually \emph{not alone} but interacts with other targets (see textual queries in Fig.~\ref{fig:task_compare} (a) and (b)). Localization of objects of interest, together with the interacting counterparts, provides richer contextual information for objects, and thus enables more comprehensive spatio-temporal understanding of the video, which greatly benefits many applications including video surveillance, robotics, sport analysis, and so on. In existing STVG, nevertheless, the crucial interacting counterparts are often neglected in localization, restricting STVG for more comprehensive analysis.

To mitigate the aforementioned limitations of current STVG, the key is to have the ability to locate \emph{every} target mentioned in the textual query, much like how humans do. 

Thus motivated, we in this paper introduce a new type of STVG task, dubbed \emph{\textbf{S}patio-\textbf{T}emporal \textbf{Omni}-Object \textbf{V}ideo-\textbf{G}rounding} (or \textbf{\emph{OmniSTVG}}). Different from existing STVG locating only a \emph{single} object, OmniSTVG aims at localizing \emph{all} targets mentioned in the given query from the video. For each object in the query, a spatio-temporal tube is predicted as the localization result. By doing so, OmniSTVG enables localization of \emph{not only} arbitrary number (\eg, one or multiple) of targets of interest \emph{but also} their interacting counterparts in the video, which simultaneously resolves two limitations of current STVG and therefore leads to more practical applications. It is worthy to notice that, OmniSTVG is a natural extension of classic STVG task, aiming to further push its frontier for more comprehensive multimodal video understanding. Concept-wise, OmniSTVG, to some extent, is inspired by the idea of \emph{segmentation anything} (SA)~\cite{kirillov2023segment}. The difference is, SA aims to segment any regions in an image, while OmniSTVG locates any mentioned objects in the textual query from an untrimmed video sequence.

In order to facilitate exploration of OmniSTVG, we propose \textbf{\emph{BOSTVG}}, a novel large-scale benchmark dedicated to spatio-temporal omni-object video grounding. More specifically, BOSTVG comprises 10,018 videos with 10.2 million frames, and covers a wide selection of 287 categories from diverse
scenarios. Each video in our BOSTVG, paired with a free-form textual query, contains a varying number of objects to locate, ranging from 1 to 10 with an average of 2.4. Each object is manually annotated with a spatio-temporal tube (\ie, a set of bounding boxes). To ensure high quality, all the tube annotations in each sequence are carefully inspected and refined when needed through multiple rounds. To the best of our knowledge, BOSTVG is to date the first and largest benchmark dedicated to OmniSTVG.

Furthermore, to encourage future research in developing OmniSTVG methods on BOSTVG, we propose a simple yet effective model named \emph{\textbf{OmniTube}}. Specifically, OmniTube is built upon current Transformer-based STVG method~\cite{yang2022tubedetr}. It comprises a multimodal encoder for video and text feature fusion and a decoder for localization. Different from current STVG models (\eg,~\cite{yang2022tubedetr,gu2024context,jin2022embracing,lin2023collaborative}) that locate only a single target, our OmniTube learns \emph{simultaneously} multiple sets of object queries in decoder to ground \emph{all} objects in the video. For improving localization, we leverage visual information in video guided by textual feature to generate queries, which benefits learning better query features for target grounding. In order to form a spatio-temporal box tube for each target, a simple strategy is designed to match detection results across different frames in the video. Despite simplicity, OmniTube shows promising results and expects to provide a reference for future research on our OmniSTVG task.
\renewcommand{\arraystretch}{1}
\begin{table*}[!t]
\setlength{\tabcolsep}{6pt}
  \centering
  \caption{Summary of BOSTVG and comparison to other benchmarks. SO: Single-Object; MO: Multi-Object; AO: All-Object. Please note that, since DVD-ST is not released at this moment, we report its statistics available in the original paper~\cite{ji2024described} for comparison.}\vspace{-2mm}
  \resizebox{0.99\textwidth}{!}{
    \begin{tabular}{rcccccccccccc}
    \Xhline{1.2pt}
    \rowcolor[HTML]{d0e6f7} Benchmark & Year  & Videos & \tabincell{c}{Object\\classes} & \tabincell{c}{Mean\\frames} & \tabincell{c}{Total\\frames} & \tabincell{c}{Total \\duration} & \tabincell{c}{Min\\obj.} & \tabincell{c}{Mean\\ obj.} & \tabincell{c}{Max \\obj.} & \tabincell{c}{Total\\obj.} & \tabincell{c}{Num. of \\queries} & \tabincell{c}{Dataset\\focus} \\
    \hline\hline
    \textbf{STPR}~\cite{yamaguchi2017spatio}  & 2017  & 5,293  & 1     & 260   & 1.4M  & 14 hours    & 1     & 1.0   & 1     & 5,828 & 30,365 & SO \\
    \textbf{VID-Sentence}~\cite{ChenMLW19}  & 2019  & 5,318  &  30    &  294  & 2.3M  &  21 hours   &   1   &  1.0  &  1    & 7,654 & 7,654 & SO \\
    \textbf{VidSTG}~\cite{zhang2020does} & 2020  & 6,924  & 79    & 798   & 5.5M  & 53 hours    & 1     & 1.0     & 1     & 6,924  & 99,943 & SO \\
    \textbf{HCSTVG-v1}~\cite{tang2021human} & 2021  & 5,660  & 1     & 522   & 3.0M  & 31 hours    & 1     & 1.0     & 1     & 5,660  & 5,660  & SO \\
    \textbf{HCSTVG-v2}~\cite{tang2021human} & 2021  & 16,544 & 1     & 522   & 8.6M  & 92 hours   & 1     & 1.0     & 1     & 16,544 & 16,544 & SO \\
    \textbf{DVD-ST}~\cite{ji2024described} & 2024  & 2,750  & 163   & -     & -     & -     & 0     & 1.8   & 12    & 4,950     & 5,734  & SO, MO \\
    \hline
    \rowcolor[HTML]{e9f7ef} \textbf{BOSTVG} (ours) & 2025  & 10,018 & 287   & 1,014  & 10.2M & 102 hours  & 1     & 2.4     & 10    & 24,175 & 10,018 & AO  \\
    \Xhline{1.2pt}
    \end{tabular}}
  \label{tab:addlabel}\vspace{-3mm}
\end{table*}%

We notice a concurrent work~\cite{ji2024described} which has similar spirit with this work by supporting grounding multiples in videos. Compared to~\cite{ji2024described}, our work mainly \emph{differs} in three aspects. First, \emph{concept-wise}, OmniSTVG grounds \emph{all} objects mentioned in query, while the work of~\cite{ji2024described} localizes only \emph{partial} targets, leading to limitations in comprehensive understanding. Second, \emph{task-} and \emph{method-wise}, the work of~\cite{ji2024described} locates only objects of the \emph{same} class, while OmniSTVG and OmniTube enable localization of objects of \emph{different} categories, making it more flexible and practical. Third, \emph{dataset-wise}, BOSTVG contains 10,018 videos, which is much larger than dataset in~\cite{ji2024described} with 2,750 videos.

In summary, we make the following contributions: \ding{171} We introduce OmniSTVG, a new STVG task that locates all objects mentioned in the query toward more flexible and comprehensive understanding; \ding{170} We present BOSTVG, a large-scale dataset with 10,018 videos and more than 10 million frames from 287 categories for OmniSTVG; \ding{168} We propose OmniTube, a simple but effective method to facilitate future research of OmniSTVG; \ding{169} We demonstrate that OmniTube achieves promising performance for OmniSTVG, aiming to offer a reference and provide guidance for future research.

\section{Related Work}

\textbf{STVG Benchmarks.} Benchmarks are important for facilitating the development of STVG. The work of~\cite{yamaguchi2017spatio} proposes the STPR for spatially and temporally grounding pedestrians from trimmed videos. Similarly, VID-Sentence~\cite{ChenMLW19} is introduced also for object grounding within trimmed videos, but comprises more categories. For a more practical setting of STVG, HCSTVG-v1~\cite{tang2021human} is presented to locate human objects in the untrimmed videos, making it more challenging. Later, HCSTVG-v2~\cite{tang2021human} is introduced via expanding from HCSTVG-v1 using extra videos. Different from HCSTVG-v1/v2, VidSTG~\cite{zhang2020does}, collected from VidOR~\cite{shang2019annotating} for object relation detection in a video, aims at spatio-temporal video grounding from both declarative and interrogative sentences. In addition, besides human category, VidSTG also provides other object classes in the query and video for localization, aiming at generic STVG. Unlike the above benchmarks only for single-target localization, the recently introduced DVD-ST in~\cite{ji2024described} provides a platform for grounding multiple targets while ignoring interacting counterparts in localization. 

\textbf{\emph{Different}} from all the aforementioned benchmarks, our BOSTVG is specially developed for a new STVG task, OmniSTVG, which aims to ground \emph{all} target objects mentioned in the textual query. Therefore, in our BOSTVG, each target in the query is annotated with a spatio-temporal tube  (see Fig.~\ref{fig:task_compare} (b) again for annotation examples in BOSTVG). 

\vspace{0.3em}
\noindent
\textbf{STVG Algorithms.} STVG algorithms have witnessed great progress recently. Early approaches (\eg,~\cite{tang2021human,ZhangZLHY20,zhang2020does}) typically adopt a two-stage pipeline, which first detects candidate region proposals with a pre-trained detector (\eg,~\cite{RenHGS15}) and then finds correct region proposals with an extra model. Despite straightforwardness, these two-stage methods heavily rely on the pre-trained detection model, and their performance is thus restricted by the capacity of the used detector. In order to overcome this limitation, recent STVG methods (\eg,~\cite{su2021stvgbert,gu2024context,yang2022tubedetr,jin2022embracing,lin2023collaborative,wasim2024videogrounding,gu2025knowing}), inspired by DETR~\cite{carion2020end}, switch to a one-stage design directly predicting a spatial-temporal tube for localization using Transformer~\cite{VaswaniSPUJGKP17}, without adopting any external detectors. Compared to two-stage methods, such one-stage framework shows superior performance for its end-to-end training pipeline. Our OmniTube is also a one-stage Transformer-based approach. Nonetheless, \textbf{\emph{different}} from the aforementioned methods that are designed for localizing only a \emph{single} target from the video, our OmniTube aims to locate \emph{all} objects mentioned in the query for more comprehensive multimodal video understanding.

\vspace{0.3em}
\noindent
\textbf{Video Grounding.} Video grounding aims to localize video content provided a custom query. Besides STVG, there exist many other video grounding tasks, such as moment retrieval (\eg,~\cite{gao2017tall,lei2020tvr,anne2017localizing,mun2020local,zeng2020dense}), query-based video summarization (\eg,~\cite{sharghi2017query,wu2022intentvizor}, video highlight detection (\eg,~\cite{gygli2016video2gif,hong2020mini,badamdorj2021joint,xu2021cross}), etc. \textbf{\emph{Unlike}} these tasks for only temporal grounding in the video, OmniSTVG and STVG aim at both spatial and temporal grounding from videos. Particularly, our OmniSTVG seeks to localize all mentioned targets in the query, making it more challenging yet practical in applications.

\begin{figure*}[t!]
    \centering
    \begin{minipage}[b]{0.32\linewidth}
        \includegraphics[width=1\linewidth]{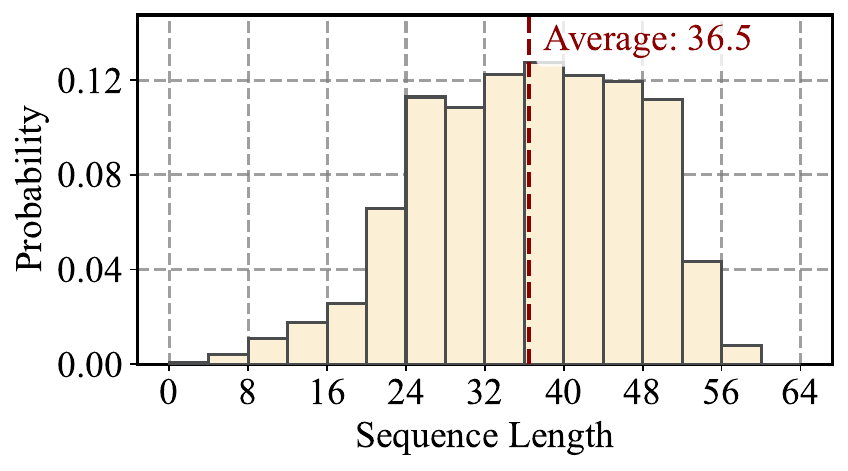}
        \captionsetup{labelformat=empty, font={scriptsize}, skip=2pt}
        \subcaption*{(a) Distribution of video length (in seconds)}
        \label{subfig:tracj length}
    \end{minipage}
    \begin{minipage}[b]{0.32\linewidth}
        \includegraphics[width=1\linewidth]{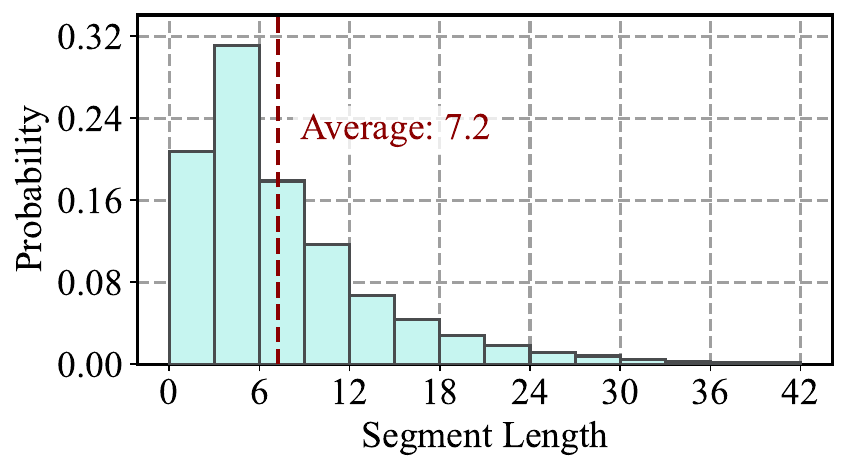}
        \captionsetup{labelformat=empty, font={scriptsize}, skip=2pt}
        \subcaption*{(b) Distribution of segment length (in seconds)}
        \label{subfig:seq length}
    \end{minipage}
    \begin{minipage}[b]{0.32\linewidth}
        \includegraphics[width=1\linewidth]{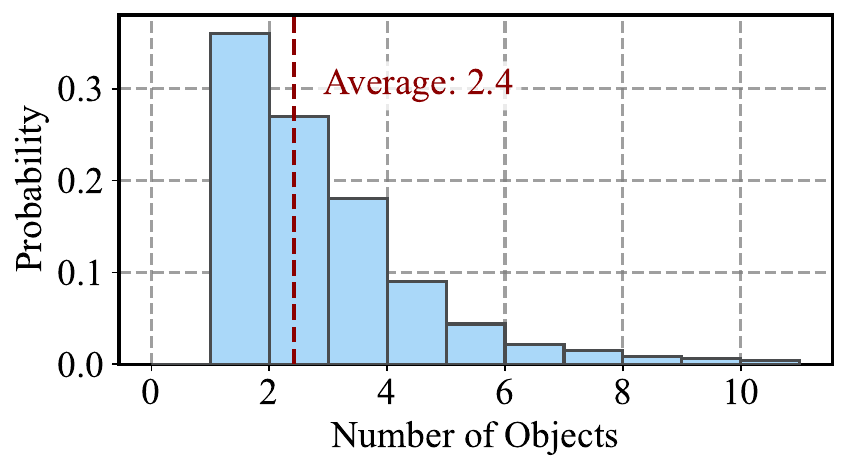}
        \captionsetup{labelformat=empty, font={scriptsize}, skip=2pt}
        \subcaption*{(c) Distribution of number of obejcts}
        \label{subfig:relation num}
    \end{minipage}
    \vspace{-2mm}
    \caption{Representative statistics on BOSTVG, including distributions of video length (in seconds) in image (a), temporal segment length (in seconds) in image (b), and the number of target objects for grounding in image (c).}
    \label{fig:statistic}
    \vspace{-4mm}
\end{figure*}

\begin{figure}[!t]
    \centering
    \includegraphics[width=0.9\linewidth]{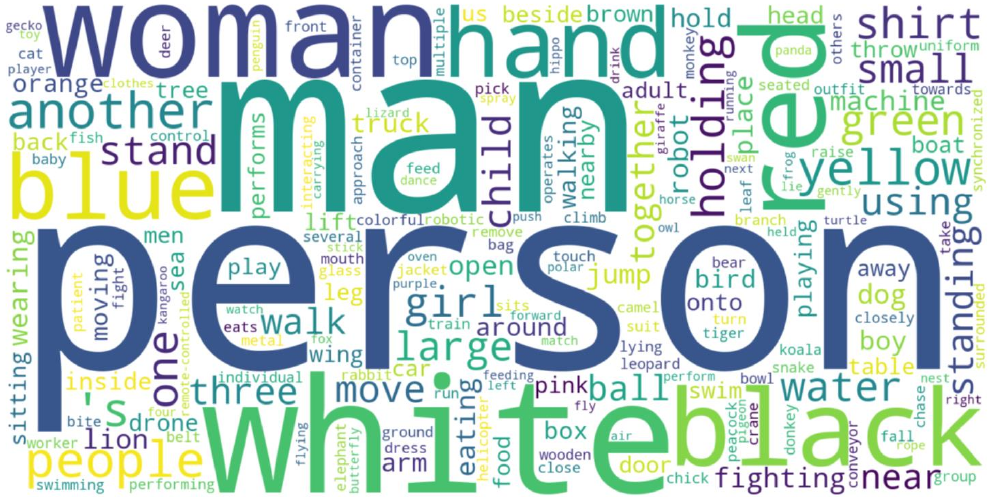}\vspace{-1mm}
    \caption{Wordcloud of all textual queries.}
    \label{fig:word}\vspace{-4mm}
\end{figure}

\section{The Proposed BOSTVG}

\subsection{Design Principle}

BOSTVG aims to offer a dedicated platform for facilitating development of OmniSTVG. For such a purpose, we follow the principles below in developing BOSTVG:

\begin{itemize}
    \item \emph{\textbf{Dedicated benchmark.}} The motivation of BOSTVG is to provide a novel benchmark dedicated to OmniSTVG. The video and its paired textual query are required to comprise a varying number of target objects (\eg, one or multiple) to localize, aligning with the goal of OmniSTVG.

    \item \emph{\textbf{Large scale.}} Developing deep learning-based models for OmniSTVG requires abundant training samples. Besides training, a real system needs evaluation on various cases. Thus, we expect BOSTVG to contain at least 10K videos with each corresponding to a textual query, which benefits both large-scale training and assessment of deep models.
    
    \item \textit{\textbf{Diverse object classes.}} An important aim of BOSTVG is to facilitate the development of general OmniSTVG models that can locate targets from different classes. To this end, the new dataset expects to contain at least 200 categories, collected from various scenarios, for grounding.
    
    \item \emph{\textbf{High quality.}} High-quality annotations are essential for a benchmark in both training and evaluation. To ensure the high quality, each video of BOSTVG is manually labeled with precise spatial-temporal box tubes through multiple rounds of inspections and refinements. 
\end{itemize}

\subsection{Data Acquisition}

BOSTVG aims to foster general and comprehensive spatial-temporal video grounding by containing rich object classes from diverse scenarios. To this end, 287 object classes that are appropriate for OmniSTVG are selected in BOSTVG. These categories are chosen from different sources, mainly including ImageNet~\cite{deng2009imagenet} and V3Det~\cite{wang2023v3det}, and organized in a coarse-to-fine hierarchical structure. Due to limited space, we show detailed classes in the \emph{supplementary material}.

After determining all object categories in BOSTVG, we then search for raw videos of each class under various scenarios from YouTube, currently the largest and most popular video platform with many real-world videos. All videos are collected under the Creative Commons License and used for research purpose only. Initially, we have collected over 15K videos using keywords aligned with the object classes. Then, we conduct careful inspections on each video to verify its suitability for OmniSTVG. Specifically, if there is at least one video clip suitable for our task, we keep this video; otherwise, we discard the video. This process is carried out by our experts (\eg, students working on related field). For the qualified video sequences, we select one clip from each of them. Eventually, we gather 10,018 videos for BOSTVG, with each provided a textual query by our experts.

Finally, we create a large-scale dataset, called BOSTVG, for OmniSTVG. BOSTVG covers 287 classes and contains 10,018 videos with 10.2 million frames from diverse scenarios. Its average video length is 1,014 frames. Each video contains a varying number of targets ranging from 1 to 10. It is worthy to notice that, we do not consider the case of \emph{none} target in the video in our work and focus on localizing targets that appear within the video. Tab.~1 summarizes our BOSTVG and its comparison to classic STVG benchmarks.

\renewcommand{\arraystretch}{1.0}
\begin{table}[!t]
  \centering
  \caption{Comparison between training and testing sets.}\vspace{-2mm}
  \resizebox{0.46\textwidth}{!}{
    \begin{tabular}{cccccc}
    \Xhline{1.2pt}
    \rowcolor[HTML]{d0e6f7}  & Videos & \tabincell{c}{Mean\\frames} & \tabincell{c}{Total\\frames} & \tabincell{c}{Mean\\ obj.} &\tabincell{c}{Total\\obj.} \\
    \hline\hline
    BOSTVG$_\text{Tra}$   & 8,106    &  1,014  & 8.22M  &   2.4      &  19,567   \\
    BOSTVG$_\text{Tst}$   & 1,912    &  1,015  & 1.94M &       2.4 &   4,608  \\
    \Xhline{1.2pt}
    \end{tabular}}
  \label{tab:splitset}\vspace{-4mm}
\end{table}%

\subsection{Data Annotation}

In BOSTVG, each video is offered two types of annotations based on textual query  generated by experts during data collection, comprising the start and end timestamps for temporal localization and bounding box tubes of objects for spatial localization. Specifically, given the pair of video and textual query, we first identify a temporal segment in the video that corresponds to the description in query, and mark it with start and end timestamps. Afterwards, we label each object mentioned in query with a consistent spatio-temporal tube formed by a set of boxes in each frame of temporal segment.

\begin{figure*}[!t]
    \centering
    \includegraphics[width=0.93\linewidth]{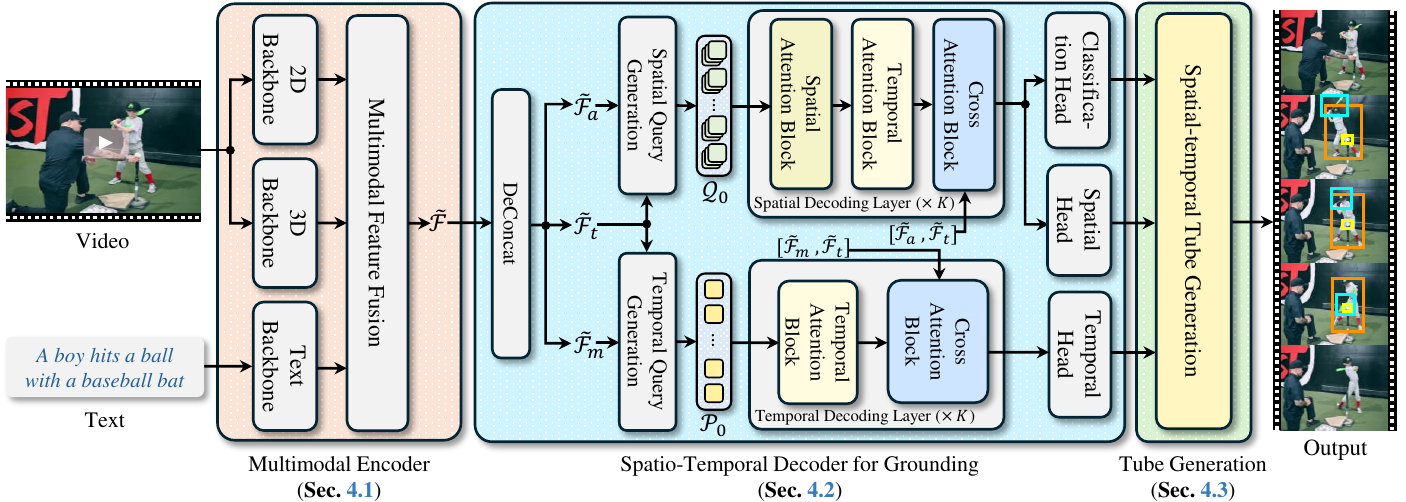}\vspace{-2mm}
    \caption{Overview of the proposed OmniTube, which consists of a multimodal encoder, a spatio-temporal decoder, and a spatial-temporal box tube generation module to localize all mentioned target objects in the textual query for OmniSTVG.}
    \label{fig:mainnetwork}\vspace{-4mm}
\end{figure*}

To ensure high-quality annotations of BOSTVG, we use a multi-round strategy. Specifically, a few experts who work on related problems first manually annotate the start and end timestamps for each sequence. Then, the temporal segment of each video, marked by start and end timestamps, is manually labeled by an annotation team that is formed by a few volunteers and an expert. After this initial round, the spatio-temporal tube annotations will be sent to a validation team formed by three experts for inspection. If the initial annotations are not unanimously agreed by all experts, they will be returned back to the original
labeling team for refinement. We repeat this inspection-refinement process until the annotations of all videos are qualified. Due to limited space, we show the pipeline of our annotation process in \emph{supplementary material}. Fig.~\ref{fig:task_compare} displays a few annotation examples in BOSTVG, and more can be seen in \emph{supplementary material}.

\vspace{0.2em}
\noindent
\textbf{Annotation accuracy analysis.} To analyze the accuracy of our annotation, we randomly select 100 videos of BOSTVG and ask an independent group of external experts to inspect and re-label them. Then, we compute the Intersection over Union (IoU) of new and original spatio-temporal tubes. The IoU for these selected sequences is 0.90, which validates the accuracy as well as reliability of our annotations.

\vspace{0.2em}
\noindent
\textbf{Statistics of annotation.} To better understand BOSTVG, we display
some statics in Fig.~\ref{fig:statistic}, including distributions of video length, temporal segment length, and number of objects. From Fig.~\ref{fig:statistic} (c), we can see that most videos contain 1 to 6 objects, which is close to real scenarios and makes BOSTVG more suitable for practical applications.

\subsection{Dataset Split and Evaluation Metric}

\textbf{Dataset Split.} BOSTVG contains a total of 10,018 videos. Among them, 8,106 videos are selected for the training set, dubbed BOSTVG$_{\text{Tra}}$, and the rest 1,912 sequences are used for the testing set, dubbed BOSTVG$_{\text{Tst}}$. Tab.~\ref{tab:splitset} displays the comparison of training and test sets. It is worthy to notice that, in split, we try our best to keep distributions of training and test sets similar. For both training and testing sets, each video is paired with a textual involving a varying number of targets for grounding, meeting the aim of OmniSTVG.

In addition, to enable in-depth analysis, we further divide the test set into three subsets based on the number of targets in the video, including BOSTVG$_{\text{Tst}}$-Low with 1-3 targets in each video (1,566 samples), BOSTVG$_{\text{Tst}}$-Medium with 4-6 targets in each video (273 samples), and BOSTVG$_{\text{Tst}}$-High with more than 7 targets in each video (73 samples). 

\vspace{0.2em}
\noindent
\textbf{Evaluation Metric.} Following current benchmarks~\cite{ChenMLW19,zhang2020does}, we utilize multiple metrics, including \emph{m\_tIoU}, \emph{m\_vIoU}, and \emph{vIoU@R} for evaluation. Please note, the calculation of m\_vIoU and vIoU@R here needs to consider all spatial-temporal box tubes in the video, instead of a single one as in other benchmarks, because the video in our BOSTVG may contain multiple objects. Due to limited space, please kindly refer to the \emph{supplementary material} for the detailed formulations of these metrics.

\section{OmniTube: A New Baseline for OmniSTVG}

\textbf{Overview.} We propose OmniTube, a new baseline specially designed for OmniSTVG. Similar to current STVG models (\eg,~\cite{yang2022tubedetr,jin2022embracing,lin2023collaborative}), inspired by DETR~\cite{carion2020end}, OmniTube adopts an encoder-decoder architecture. As shown in Fig.~\ref{fig:mainnetwork}, OmniTube mainly consists of three components, including a multimodal encoder for feature extraction and fusion (Sec.~\ref{encoder}), a spatio-temporal decoder for target object position learning (Sec.~\ref{decoder}), and a box tube generation module (Sec.~\ref{tube}), to localize all mentioned objects in the text for OmniSTVG.

\subsection{Multimodal Encoder}
\label{encoder}

Given a video and the text, the multimodal encoder first extracts their features and then performs multimodal feature fusion as described in the following.

\vspace{0.3em}
\noindent
\textbf{Feature Extraction.} Given the video sequence, we extract both its 2D appearance and 3D motion features. Specifically, we first sample $N_v$ frames from the video, and then adopt ResNet-101~\cite{he2016deep} and VidSwin~\cite{liu2022video} for appearance and motion feature extraction, respectively. The appearance feature is represented as  $\mathcal{F}_{a}=\{f^{a}_i\}_{i=1}^{N_v}$, where $f^{a}_i \in \mathbb{R}^{H \times W \times D_a}$ with $H$, $W$ and $D_a$ the height, width, and channel dimensions, and the motion feature as $\mathcal{F}_{m}=\{f^{m}_i\}_{i=1}^{N_v}$, where $f^{m}_i \in \mathbb{R}^{H \times W \times D_m}$ with $D_m$ the channel dimension. 

For the text, we use RoBERTa~\cite{RoBERTa} for feature extraction. We tokenize it to a sequence with $N_t$ words, and then apply RoBERTa on the sequence to produce textual feature $\mathcal{F}_{t} = \{f^t_i\}_{i=1}^{N_t}$, where $f^t_i \in \mathbb{R}^{D_t}$ with $D_t$ the feature channel.

\vspace{0.3em}
\noindent
\textbf{Multimodal Feature Fusion.}
We generate the multimodal feature by fusing appearance, motion, and textual features. 
Similar to~\cite{gu2024context,gu2025knowing}, we first project them to the same channel dimension $D$ and then concatenate them to obtain the initial multimodal feature $\mathcal{F}=\{f_i\}_{i=1}^{N_v}$ as follows,
\begin{equation}\nonumber
\setlength{\abovedisplayskip}{5pt}
\setlength{\belowdisplayskip}{5pt}
f_i = [\underbrace{f^a_{i_1},..., f^a_{i_{H \times W}}}_{\text{app. feature $f^a_i$}},
\underbrace{f^m_{i_1},..., f^m_{i_{H \times W}}}_{\text{motion feature $f^m_i$}}, \underbrace{f_{1}^{t},..., f_{N_t}^{t}}_{\text{text feature $f_{i}^{t}$}}]
\end{equation}
where $f_i$ is the multimodal feature in frame $i$. Then, we fuse the features using self-attention encoder~\cite{waswani2017attention} to generate the multimodal feature $\mathcal{\tilde{F}}$ as follows,
\begin{equation}\nonumber
\setlength{\abovedisplayskip}{5pt}
\setlength{\belowdisplayskip}{5pt}
\mathcal{\tilde{F}}=\mathtt{SAEncoder}(\mathcal{F}+\mathcal{E}_{\text{pos}}+\mathcal{E}_{\text{type}})
\end{equation}
where $\mathcal{E}_{\text{pos}}$ and $\mathcal{E}_{\text{type}}$ denote position and type embeddings. $\mathtt{SAEncoder}(\cdot)$ is self-attention encoder with $L$ ($L$=6) standard self-attention encoder blocks. Due to space limitation, please see its architecture in \emph{supplementary material}.

\subsection{Spatio-Temporal Decoder for Grounding}
\label{decoder}

To obtain target position information from multimodal feature $\mathcal{\tilde{F}}$, we design a spatio-temporal decoder composed of a spatial decoder and a temporal decoder. The former learns spatial information for all objects in the text, while the later aims to obtain the temporal information for grounding.

\subsubsection{Spatial Omni-Object Decoder for Grounding}

\textbf{Spatial Query Generation.} Unlike current STVG methods locating a single target, OmniTube localizes all objects in text, and thus introduces multiple queries for each frame. To explore target cue for better localization, we leverage text as guidance to select target-relevant features in video for generating object queries. To this end, we first extract features from $\tilde{\mathcal{F}}$ by deconcatenation $[\tilde{\mathcal{F}}_a, \tilde{\mathcal{F}}_m, \tilde{\mathcal{F}}_t]$=$\mathtt{DeConcat}(\tilde{\mathcal{F}})$, where $\tilde{\mathcal{F}}_a$/$\tilde{\mathcal{F}}_m$/$\tilde{\mathcal{F}}_t$ are appearance/motion/textual features. Then, we utilize appearance and textual features to generate the spatial queries. Specifically, we first average textual feature $\tilde{\mathcal{F}}_t$ via $\bar{\mathcal{F}}_t$=$\mathtt{Avg}(\tilde{\mathcal{F}}_t)$. Then, we calculate similarity between $\bar{\mathcal{F}}_t$ and $\tilde{\mathcal{F}}_a$, and adopt the $M$ most similar features to generate initial query $\mathcal{Q}_{0}$ by average pooling, as follows, 
\begin{equation}\nonumber
\setlength{\abovedisplayskip}{5pt}
\setlength{\belowdisplayskip}{5pt}
\begin{split}
    \mathcal{Q}_{0} &= \{\{q_{i,j}^{0}\}_{j=1}^{N_q}\}_{i=1}^{N_v}, \\ q_{i,j}^{0}& =\mathtt{AvgPooling}(\mathtt{Top}_{\mathtt{M}}(\mathtt{Sim}(\tilde{\mathcal{F}}_a,\bar{\mathcal{F}}_t)))
\end{split}
\end{equation}
where $q_{i,j}^{0}$ is the feature of the $j^{\text{th}}$ query in frame $i$, and $N_q$ the number of queries per frame. $\mathtt{Sim()}$ and $\mathtt{Top}_{\mathtt{M}}()$ are the operations to calculate similarities and to pick up top $M$ elements, respectively. Compared with previous approaches, exploration of target specific cues for generating queries can effectively improve localization as in our experiments.

\vspace{0.3em}
\noindent
\textbf{Spatial Decoding.}
After generating the $\mathcal{Q}_{0}$, we fed it to the spatial (omni-object) decoder with $K$ ($K$=$6$) layers for interaction with the multimodal feature. To enhance queries, in each layer we design two simple yet effective spatial and temporal attention blocks to capture their spatial and temporal relation before interacting with the multimodal feature.  

Specifically, let $\mathcal{Q}_{k-1}$ represent query features sent to the $k^{\text{th}}$ ($1\le k \le K$) layer for learning $\mathcal{Q}_{k}$. We first perform spatial attention on queries of the same frame, as follows, 
\begin{equation}\nonumber
\setlength{\abovedisplayskip}{5pt}
\setlength{\belowdisplayskip}{5pt}
\{\hat{q}_{i,j}^{k-1}\}_{j=1}^{N_q} = \mathtt{SABlock}(\{q_{i,j}^{k-1}\}_{j=1}^{N_q}) \;\;\; i=1,2,\cdots,N_v
\end{equation}
where $\hat{\mathcal{Q}}_{k-1}$=$\{\{\hat{q}_{i,j}^{k-1}\}_{j=1}^{N_q}\}_{i=1}^{N_v}$ denotes the query features after spatial attention. $\mathtt{SABlock}(\cdot)$ is spatial attention block implemented with self-attention as shown in \emph{supplementary material}. After this, to further capture the temporal relation, we apply the temporal attention on the query features of the same object across different frames, as follows,
\begin{equation}\nonumber
\setlength{\abovedisplayskip}{5pt}
\setlength{\belowdisplayskip}{5pt}
\{\tilde{q}_{i,j}^{k-1}\}_{j=1}^{N_q} = \mathtt{TABlock}(\{\hat{q}_{i,j}^{k-1}\}_{i=1}^{N_v}) \;\;\; j=1,2,\cdots,N_q
\end{equation}
where $\tilde{\mathcal{Q}}_{k-1}$=$\{\{\tilde{q}_{i,j}^{k-1}\}_{j=1}^{N_q}\}_{i=1}^{N_v}$ represents query features after the temporal attention block $\mathtt{TABlock}(\cdot)$ implemented with self-attention as in \emph{supplementary material}. 

Next, we learn the spatial position of objects by interacting queries with multimodal feature. In OmniTube, spatial localization leverages the appearance and text features. Specifically, we interact $\tilde{\mathcal{Q}}_{k-1}$ with the multimodal feature via cross-attention for learning $\tilde{\mathcal{Q}}_{k}$, as follows,
\begin{equation}\nonumber
\setlength{\abovedisplayskip}{5pt}
\setlength{\belowdisplayskip}{5pt}
    \mathcal{Q}_{k} = \mathtt{CrossAttBlock}(\mathcal{\tilde{Q}}_{k-1}, [\mathcal{\tilde{F}}_a, \mathcal{\tilde{F}}_t])
\end{equation}
where $\mathtt{CrossAttBlock}(\textbf{z},\textbf{u})$ represents the cross-attention block with $\textbf{z}$ generating query and $\textbf{u}$ key/value.

Finally, with $\mathcal{Q}_{K}$ after the $K^{\text{th}}$ layer in decoder, we adopt a spatial head, consisting of an MLP module, to predict the bounding boxes of the targets via $\mathcal{B} = \mathtt{SpatialHead}(\mathcal{Q}_K)$, where $\mathcal{B} \in \mathbb{R}^{N_v \times N_q \times D_b}$ and $D_b=4$ is the central position, width and height of predicted box.
In addition, inspired by MDETR~\cite{kamath2021mdetr}, we predict the index for each bounding box, which corresponds to positional index of words in the original text, and is used to determine the class of each bounding box, via $\mathcal{G} = \mathtt{ClsHead}(\mathcal{Q}_K)$, 
where $\mathtt{ClsHead}(\cdot)$ is a MLP module and $\mathcal{G} \in \mathbb{R}^{N_v \times N_q \times N_t}$ with $N_t$ denoting the maximum positional indexes for any given sentence.

\subsubsection{Temporal Decoder for Grounding.}

\textbf{Temporal Query Generation.} Temporal decoder predicts start and end timestamps. Similar to spatial decoder, we use target cues for temporal query generation. Specifically, we leverage target-relevant motion features selected by the textual features to produce the initial query $\mathcal{P}_{0}$ as follows,
\begin{equation}\nonumber
\setlength{\abovedisplayskip}{5pt}
\setlength{\belowdisplayskip}{5pt}
\begin{split}
    \mathcal{P}_{0} = \{p_{i}^{0}\}_{i=1}^{N_v}, \;\; p_{i}^{0} =\mathtt{AvgPooling}(\mathtt{Top}_{\texttt{M}}(\mathtt{Sim}(\tilde{\mathcal{F}}_m,\bar{\mathcal{F}}_t)))
\end{split}
\end{equation}
where $p_{i}^{0}$ is query feature in frame $i$. $\bar{\mathcal{F}}_t$ is the pooled textual feature and $\tilde{\mathcal{F}}_m$ the motion feature extracted from $\tilde{\mathcal{F}}$. It is worth noting that, in OmniSTVG, all targets share the same start and end timestamps with the textual expression. Thus, each frame $i$ is assigned with a single initial query $p_i^0$.

\vspace{0.3em}
\noindent
\textbf{Temporal Decoding.} In temporal decoding, we send $\mathcal{P}_{0}$ to a decoder with $K$ layers for interaction with multimodal feature. Specifically, let $\mathcal{P}_{k-1}$=$\{p_{i}^{k-1}\}_{i=1}^{N_v}$ be query features fed to the $k^{\text{th}}$ ($1\le k \le K$) layer for learning $\mathcal{P}_{k}$, where $p_{i}^{k-1}$ is the feature of frame $i$. To capture temporal relation, we first perform temporal attention on $\mathcal{P}_{k-1}$, as follows,
\begin{equation}\nonumber
\setlength{\abovedisplayskip}{5pt}
\setlength{\belowdisplayskip}{5pt}
\{\tilde{p}_{i}^{k-1}\}_{i=1}^{N_v} = \mathtt{TABlock}(\{p_{i}^{k-1}\}_{i=1}^{N_v}) 
\end{equation}
where $\tilde{\mathcal{P}}_{k-1}$=$\{\tilde{p}_{i}^{k-1}\}_{i=1}^{N_v}$ is the query feature after temporal attention. After this, we interact $\tilde{\mathcal{P}}_{k-1}$ with the multimodal feature using cross-attention for learning $\mathcal{P}_{k}$, as follows,
\begin{equation}\nonumber
\setlength{\abovedisplayskip}{5pt}
\setlength{\belowdisplayskip}{5pt}
    \mathcal{P}_{k} = \mathtt{CrossAttBlock}(\mathcal{\tilde{P}}_{k-1}, [\mathcal{\tilde{F}}_m, \mathcal{\tilde{F}}_t])
\end{equation}
where $\mathcal{\tilde{F}}_m$ and $\mathcal{\tilde{F}}_t$ are motion and textual features from $\tilde{\mathcal{F}}$.

After $K$ layers in temporal decoder, we can obtain $\mathcal{P}_{K}$ and employ a temporal head that is implemented by an MLP module to predict the start and end timestamps through $\mathcal{H} = \mathtt{TemporalHead}(\mathcal{P}_{K})$, where $\mathcal{H} \in \mathbb{R}^{N_v \times 2}$ contains the start probabilities $\mathcal{H}_s$ and end probabilities $\mathcal{H}_e$ of $N_v$ frames for the temporal localization of targets.

\subsection{Spatial-temporal Tube Generate}
\label{tube}

To generate the spatial-temporal tube for each target object, we first use tubelet matching to connect the bounding boxes across frames, then filter tubelets using class information.

\vspace{0.2em}
\noindent
\textbf{Tubelet Matching.}
Spatial grounding predicts $N_q$ bounding boxes for each frame. To match these boxes across different frames, we apply Hungarian matching~\cite{kuhn1955hungarian}, which is based on the spatial positions and object class of the bounding boxes, ultimately generating $N_q$ initial tubelets.

\vspace{0.2em}
\noindent
\textbf{Tubelet Filtering.} To further optimize tubelets, we first refine their temporal boundaries using start and end timestamps predicted in temporal grounding. For each tubelet, we average the class probabilities of all its bounding boxes for determining the class. After that, we remove the tubelets whose classes are not present in the text.

\subsection{Optimization}

Given a video and its textual expression, OmniTube predicts two types of localization, comprising (1) the spatial position $\mathcal{B}$ and class $\mathcal{G}$ of the bounding box in the spatial grounding, and (2) the start timestamps $\mathcal{H}_s$ and end timestamps $\mathcal{H}_e$ in the temporal grounding.
During training, given the ground truth for the spatial location $\mathcal{B}^{*}$ and class $\mathcal{G}^{*}$ of the bounding box, as well as the start timestamps $\mathcal{H}_s^{*}$ and end timestamps, we can calculate the total loss  as
\begin{equation}\nonumber
\setlength{\abovedisplayskip}{5pt}
\setlength{\belowdisplayskip}{5pt}
    \begin{split}
        \mathcal{L} =& \lambda_h \mathcal{L}_{h}((\mathcal{B}, \mathcal{G}), (\mathcal{B}^{*}, \mathcal{G}^{*}))
        +\lambda_k (\mathcal{L}_{{\text{k}}}(\mathcal{H}_s^*,\mathcal{H}_s) \\
        &+ \mathcal{L}_{{\text{k}}}(\mathcal{H}_e^*,\mathcal{H}_e))
    \end{split}
\end{equation}
where $\mathcal{L}_{h}$ denotes the Hungarian loss as in~\cite{carion2020end} (please refer to our \emph{supplementary material} for details) and $\mathcal{L}_{k}$ represents the KL divergence loss. The parameters $\lambda_h$ and $\lambda_k$ are used to balance the loss.

\begin{table}[!t]
\setlength{\tabcolsep}{5pt}
	\centering
	\renewcommand{\arraystretch}{1.0}
        \caption{Comparison with current STVG approaches on BOSTVG test set. ${\dagger}$: the algorithm is adapted for OmniSTVG with minimum modifications for input/output and trained on BOSTVG.} \vspace{-2mm}
	\scalebox{0.84}{
		\begin{tabular}{rcccc}
			\specialrule{1.5pt}{0pt}{0pt}
			\rowcolor[HTML]{d0e6f7}
			\textbf{Methods} & \textbf{m\_tIoU} & \textbf{m\_vIoU} & \textbf{vIoU@0.3} & \textbf{vIoU@0.5} \\
			\hline\hline
                \rowcolor[HTML]{eaf4fc} 
                \multicolumn{5}{c}{\centering \emph{(a)  BOSTVG$_{\text{Tst}}$-Low}} \\
		TubeDETR${\dagger}$~\cite{yang2022tubedetr} & 31.20 & 7.99 & 3.79 & 0.21 \\
            STCAT${\dagger}$~\cite{jin2022embracing} & 33.68 & 8.52 & 4.03 & 0.38\\ 
            CG-STVG${\dagger}$~\cite{gu2024context} & 32.70 & 8.29 & 4.22 & 0.32 \\ \hdashline
            Baseline (ours) & 25.54 & 5.58 & 1.34 & 0.26 \\
            OmniTube (ours) & \textbf{36.16} & \textbf{10.11} & \textbf{7.16} & \textbf{1.09} \\
		\specialrule{1.5pt}{0pt}{0pt}
            \rowcolor[HTML]{eaf4fc}  
            \multicolumn{5}{c}{\centering \emph{(b) BOSTVG$_{\text{Tst}}$-Medium}} \\
		TubeDETR${\dagger}$~\cite{yang2022tubedetr} & 30.40 & 5.81 & 0.00 & 0.00  \\
            STCAT${\dagger}$~\cite{jin2022embracing} & 31.72 & 6.20 & 0.00 & 0.00 \\ 
            CG-STVG${\dagger}$~\cite{gu2024context} & 29.70 & 5.30 & 0.37 & 0.00  \\ \hdashline
            Baseline (ours) & 26.84 & 4.89 & 0.00 & 0.00 \\
            OmniTube (ours) & \textbf{34.89} & \textbf{7.24} & \textbf{1.85} & 0.00 \\
            \specialrule{1.5pt}{0pt}{0pt}
            \rowcolor[HTML]{eaf4fc}  
            \multicolumn{5}{c}{\centering \emph{(c) BOSTVG$_{\text{Tst}}$-High}} \\
		TubeDETR${\dagger}$~\cite{yang2022tubedetr} & 30.27 & 3.91 & 0.95 & 0.00 \\
            STCAT${\dagger}$~\cite{jin2022embracing} & 31.56 & 4.36 & \textbf{1.30} & 0.00   \\ 
            CG-STVG${\dagger}$~\cite{gu2024context} & 28.09 & 3.22 & \textbf{1.30} & 0.00 \\ \hdashline
            Baseline (ours) & 25.45 & 3.22 & 0.00 & 0.00 \\
            OmniTube (ours) & \textbf{32.27} & \textbf{4.42} & \textbf{1.30} & 0.00 \\
            \specialrule{1.5pt}{0pt}{0pt}

            \rowcolor[HTML]{eaf4fc}  
            \multicolumn{5}{c}{\centering \emph{(d) BOSTVG$_{\text{Tst}}$-Full}} \\
		TubeDETR${\dagger}$~\cite{yang2022tubedetr} & 31.05 & 7.52 & 3.14 & 0.17 \\
            STCAT${\dagger}$~\cite{jin2022embracing} & 33.31 & 8.03 & 3.35 & 0.31  \\
            CG-STVG${\dagger}$~\cite{gu2024context} & 32.09 & 7.66 & 3.56 & 0.26 \\ \hdashline
            Baseline (ours) & 25.73 & 5.38 & 1.10 & 0.21 \\
            OmniTube (ours) & \textbf{35.83} & \textbf{9.47} & \textbf{6.17} & \textbf{0.89} \\
            \specialrule{1.5pt}{0pt}{0pt}
	\end{tabular}}
	\label{tab:com}
	\vspace{-10pt}
\end{table}

\section{Experiments}

\textbf{Implementation.} Our OmniTube is implemented using PyTorch~\cite{pytorch}. We adopt ResNet-101~\cite{he2016deep}, VidSwin~\cite{vidswin}, and RoBERTa~\cite{RoBERTa} to extract appearance, motion, and text features.
Following~\cite{lin2023collaborative, jin2022embracing, gu2024context}, part of the model parameters, including 2D/text backbones and multimodal encoder, are initialized using pre-trained MDETR~\cite{kamath2021mdetr}. We train OmniTube end-to-end, keeping the 3D backbone frozen while training all other parameters. During training, we use the Adam optimizer~\cite{kingma2014adam} with an initial learning rate of $1e-5$ for the backbone and $1e-4$ for the remaining modules.
Additionally, data augmentations such as random resizing and random cropping are applied to all training videos, with the shorter side resized to 320 pixels. The video length $N_v$ depends on the duration of the video, with frames extracted at FPS=2, and the text length $N_t$ is set to 30. The channel dimensions $D_a$, $D_m$, $D_t$, and $D$ are set to 2,048, 768, 768, and 256. The parameters $\lambda_h$ and $\lambda_k$ are set to 2 and 1.

\subsection{State-of-the-art Comparison}
Since there are no available approaches specially designed for our OmniSTVG task, we adapt three STVG frameworks with source codes, including TubeDETR~\cite{yang2022tubedetr}, STCAT~\cite{jin2022embracing}, and CG-STVG~\cite{gu2024context}, with minimum modifications to their input and output parts, and compare our OmniTube to these approaches on the BOSTVG$_{\text{Tst}}$. Please \textbf{\emph{note}} that, all methods in comparison are trained on BOSTVG$_{\text{Tra}}$ for fairness.

Tab.~\ref{tab:com} demonstrates the results. From Tab.~\ref{tab:com}, we can see, OmniTube outperforms TubeDETR and STCAT on all metrics in all settings. Specifically, in the full BOSTVG$_{\text{Tst}}$, OmniTube achieves 35.83\% m\_tIoU and 9.47\% m\_vIoU scores, which largely surpasses the CG-STVG with 32.09\% m\_tIoU and 7.66\% m\_vIoU scores, the STCAT with 33.31\% m\_tIoU and 8.03\% m\_vIoU scores, and the TubeDETR with 31.05\% m\_tIoU and 7.52\% m\_vIoU scores, showing superiority. Besides, compared with our baseline, which shares the similar architecture with OmniTube but without query generation module, spatial and temporal attention blocks, OmniTube obtains absolute gains of 10.10\% and 4.09\% in m\_tIoU and m\_vIoU, clearly showing the effectiveness of our approach.

\begin{table}[!t]
\setlength{\tabcolsep}{5pt}
	\centering
        \caption{Ablations of spatial decoder. SQG: spatial query generation; SAB: spatial attention block; TAB: temporal attention block.}\vspace{-3mm}
	\renewcommand{\arraystretch}{1.0}
	\scalebox{0.75}{
		\begin{tabular}{cccccccc}
			\specialrule{1.5pt}{0pt}{0pt} 
			\rowcolor[HTML]{d0e6f7}   
			& \textbf{SQG} & \textbf{SAB} & \textbf{TAB}  & \textbf{m\_tIoU} & \textbf{m\_vIoU} & \textbf{vIoU@0.3} & \textbf{vIoU@0.5}  \\
			\hline
			\hline
            \ding{182} & - & - & - & 34.33 & 8.25 & 3.66 & 0.21 \\
    	  \ding{183} & - & \checkmark & \checkmark & 34.13 & 9.00 & 5.96 & 0.94 \\
             \ding{184} & \checkmark & - & \checkmark & 34.98 & 8.89 & 4.97 & 0.68 \\
             \ding{185} & \checkmark & \checkmark & - & 35.42 & 9.15 & 4.71 & 0.63 \\
             \ding{186} & \checkmark & \checkmark & \checkmark & \textbf{35.83} & \textbf{9.47} & \textbf{6.17} & \textbf{0.89} \\
		\specialrule{1.5pt}{0pt}{0pt}
	\end{tabular}}
        \label{tab:spa}
	\vspace{-5pt}
\end{table}

\begin{table}[!t]
\setlength{\tabcolsep}{6pt}
	\centering
        \caption{Ablations of the temporal decoder. TQG: temporal query generation; TAB: temporal attention block.}\vspace{-3mm}
	\renewcommand{\arraystretch}{1.0}
	\scalebox{0.76}{
		\begin{tabular}{cccccccc}
			\specialrule{1.5pt}{0pt}{0pt} 
			\rowcolor[HTML]{d0e6f7}   
			& \textbf{TQG} & \textbf{TAB} & \textbf{m\_tIoU} & \textbf{m\_vIoU} & \textbf{vIoU@0.3} & \textbf{vIoU@0.5} \\
			\hline
			\hline
    	  \ding{182} & - & - & 26.06 & 6.66 & 3.09 & 0.47 \\
             \ding{183} & - & \checkmark & 35.00 & 8.98 & 5.54 & 0.63 \\
             \ding{184} & \checkmark & - & 26.00 & 6.82 & 3.77 & 0.47 \\
             \ding{185} & \checkmark & \checkmark & \textbf{35.83} & \textbf{9.47} & \textbf{6.17} & \textbf{0.89}  \\
		\specialrule{1.5pt}{0pt}{0pt}
	\end{tabular}}
        \label{tab:tem}
	\vspace{-10pt}
\end{table}

\subsection{Ablation Study}

\noindent
\textbf{Impact of different modules in spatial decoder.}
To study the effectiveness of different modules in spatial decoder, we conduct an ablation in Tab.~\ref{tab:spa}. As in Tab.~\ref{tab:spa}, without the spatial query generation module and spatial and temporal attention blocks, the m\_tIoU and m\_vIoU scores are 34.33\% and 8.25\% (\ding{182}). When leveraging spatial and temporal attention blocks to enhance query features, we achieve comparable m\_tIoU of 34.13\% but better m\_vIoU of 9.00\% (\ding{182} \emph{v.s.} \ding{183}). When adopting our spatial query generation module with either spatial or temporal attention blocks, both m\_tIoU and m\_vIoU scores can be improved (\ding{182} \emph{v.s.} \ding{184} and \ding{182} \emph{v.s.} \ding{185}), When applying all the modules in the spatial decoder, we achieve the best results with 35.83\% m\_tIoU and 9.47\% m\_vIoU scores (\ding{186}), showing their effectiveness.

\vspace{0.3em}
\noindent
\textbf{Impact of different modules in temporal decoder.}
To further analyze the temporal decoder, we conduct an ablation study in Tab.~\ref{tab:tem}. As in Tab.~\ref{tab:tem}, without using temporal query generation and temporal attention block, the m\_tIoU and m\_vIoU scores are 26.06\% and 6.66\%, respectively (\ding{182}). When using temporal attention block for capturing temporal relationship in the video, the m\_tIoU and m\_vIoU scores can be significantly improved to 35.00\% and 8.98\% (\ding{182} \emph{v.s.} \ding{183}), showing the importance of temporal modeling for temporal localization. When using the temporal query generation alone, we achieve the similar m\_tIoU score of 26.00\% and m\_vIoU score of 6.82\% (\ding{182} \emph{v.s.} \ding{184}). When combining the temporal query generation and temporal attention block, we obtain the best 35.83\% m\_tIoU and 9.47\% m\_vIoU scores (\ding{185}), showing their necessity for OmniTube.

\vspace{0.3em}
\noindent
\textbf{Impact of different class predictions.}
In OmniTube, instead of directly producing the bounding box class for tube generation, we predict the position index of the class in the text as the bounding box class. To compare these two approaches, we conduct an ablation in Tab.~\ref{tab:class}. From Tab.~\ref{tab:class}, we observe that the  prediction of position index performs better (\ding{182} \emph{v.s.} \ding{183}). This may be because there are $287$ classes for the boxes, making direct prediction of box class difficult, whereas predicting the position index is relatively easier.

\vspace{0.3em}
\noindent
\textbf{Impact of motion information.} Besides appearance feature, we apply motion features of the video for localization. We conduct an ablation in Tab.~\ref{tab:motion}. From the Tab.~\ref{tab:motion}, we can observe that, the use of motion features enhances the performance of OmniTube for target localization (\ding{182} \emph{v.s.} \ding{183}).

\vspace{0.3em}
\noindent
\textbf{Impact of parameter $M$ in the decoder.}
In the decoder, $M$ controls the number of video features related to the text. To explore its impact, we conduct ablations as shown in Tab.~\ref{tab:param}.
We can see that the best result is obtained when $M$ is 5 (\ding{183}). 

\begin{table}[!t]
\setlength{\tabcolsep}{6.5pt}
	\centering
        \caption{Ablations of bounding box class.}\vspace{-3mm}
	\renewcommand{\arraystretch}{1.0}
	\scalebox{0.72}{
		\begin{tabular}{cccccc}
			\specialrule{1.5pt}{0pt}{0pt} 
			\rowcolor[HTML]{d0e6f7} 
			 & \textbf{Classification} & \textbf{m\_tIoU} & \textbf{m\_vIoU} & \textbf{vIoU@0.3} &  \textbf{vIoU@0.5} \\
			\hline
			\hline
             \ding{182} & Box Class & 35.36 & 8.82 & 5.44 & 0.84 \\
             \ding{183} & Position Index & \textbf{35.83} & \textbf{9.47} & \textbf{6.17} & \textbf{0.89}  \\
		\specialrule{1.5pt}{0pt}{0pt}
	\end{tabular}}
 \label{tab:class}\vspace{-5pt}
\end{table}

\begin{table}[!t]
\setlength{\tabcolsep}{8.5pt}
	\centering
        \caption{Ablations of motion information in OmniTube.}\vspace{-3mm}
	\renewcommand{\arraystretch}{1.0}
	\scalebox{0.72}{
		\begin{tabular}{cccccc}
			\specialrule{1.5pt}{0pt}{0pt} 
			\rowcolor[HTML]{d0e6f7}
			 & \textbf{Motion} & \textbf{m\_tIoU} & \textbf{m\_vIoU} & \textbf{vIoU@0.3} &  \textbf{vIoU@0.5} \\
			\hline
			\hline
             \ding{182} & - & 35.29 & 8.88 & 5.23 & 0.63 \\
             \ding{183} & \checkmark & \textbf{35.83} & \textbf{9.47} & \textbf{6.17} & \textbf{0.89} \\
		\specialrule{1.5pt}{0pt}{0pt}
	\end{tabular}}
 \label{tab:motion}\vspace{-5pt}
\end{table}

\begin{table}[!t]
\setlength{\tabcolsep}{8pt}
	\centering
        \caption{Ablations of $M$ in spatial/temporal query generation.}\vspace{-3mm}
	\renewcommand{\arraystretch}{1.0}
	\scalebox{0.72}{
		\begin{tabular}{cccccc}
			\specialrule{1.5pt}{0pt}{0pt} 
			\rowcolor[HTML]{d0e6f7} 
			 &  & \textbf{m\_tIoU} & \textbf{m\_vIoU} & \textbf{vIoU@0.3} &  \textbf{vIoU@0.5} \\
			\hline
			\hline
             \ding{182} & $M=2$ & 35.30 & 9.14 & 5.49 & 0.78 \\
             \ding{183} & $M=5$ & \textbf{35.83} & \textbf{9.47} & \textbf{6.17} & \textbf{0.89} \\
             \ding{184} & $M=10$ & 35.45 & 9.34 & 6.33 & 0.84 \\
		\specialrule{1.5pt}{0pt}{0pt}
	\end{tabular}}
 \label{tab:param}\vspace{-10pt}
\end{table}

Due to space limitations, we display additional results and analysis in the \emph{supplementary material}.

\section{Conclusion}

In this work, we introduce a novel STVG task, OmniSTVG, aiming to locate all targets in the query. To foster research on OmniSTVG, we propose a large-scale dataset BOSTVG by providing 10,018 sequences with 10.2 million frames. To the best of our knowledge, BOSTVG is the first benchmark dedicated to OmniSTVG. Moreover, to encourage future research on BOSTVG, we present OmniTube, a simple yet highly effective method for OmniSTVG. Our extensive results and analysis show the advantages of OmniTube over other approaches. By developing BOSTVG and OmniTube, we hope to inspire more future research on OmniSTVG.

{
\small
\bibliographystyle{ieeenat_fullname}
\bibliography{main}
}

\onecolumn

\appendix

\begin{center}
\section*{\textbf{\emph{OmniSTVG}: Toward Spatio-Temporal Omni-Object Video Grounding\\---Supplementary Material---}}
\end{center}

\noindent
For a better understanding of this work, we offer additional details, analysis, and results as follows:

\vspace{0.3em}
\begin{itemize}
	\setlength{\itemsep}{1pt}
	\setlength{\parsep}{1pt}
	\setlength{\parskip}{1pt}
    
   \item \textbf{A \;\; \emph{Details of Object Categories}} \\
   We present detailed object categories in BOSTVG, along with the number of sequences in each class.

   \item \textbf{B \;\; \emph{More Statistics}} \\
   We show additional statistics on BOSTVG, offering more insights into its characteristics.

   \item \textbf{C \;\; \emph{Construction Pipeline for BOSTVG and Additional Annotation Examples}} \\
   This section introduces the detailed construction pipeline of BOSTVG and showcases more annotation examples.
    
   \item \textbf{D \;\; \emph{Detailed Architectures of Modules}} \\
   In this section, we show architectures for $\mathtt{SAEncoder(\cdot)}$,
    $\mathtt{SABlock(\cdot)}$, and $\mathtt{TABlock(\cdot)}$ in the paper.
   
   \item \textbf{E \;\; \emph{Details of Hungarian Loss}} \\
   We provide an in-depth explanation of the Hungarian loss used in our method.
   
   \item \textbf{F \;\; \emph{Details of Evaluation Metrics}} \\
   In this section, we describe the evaluation metrics used for assessing the performance of different approaches.
   
   \item \textbf{G \;\; \emph{Qualitative Results}} \\
   This section shows qualitative results.
\end{itemize}

\begin{figure*}[thb]
    \centering
\includegraphics[width=0.75\linewidth]{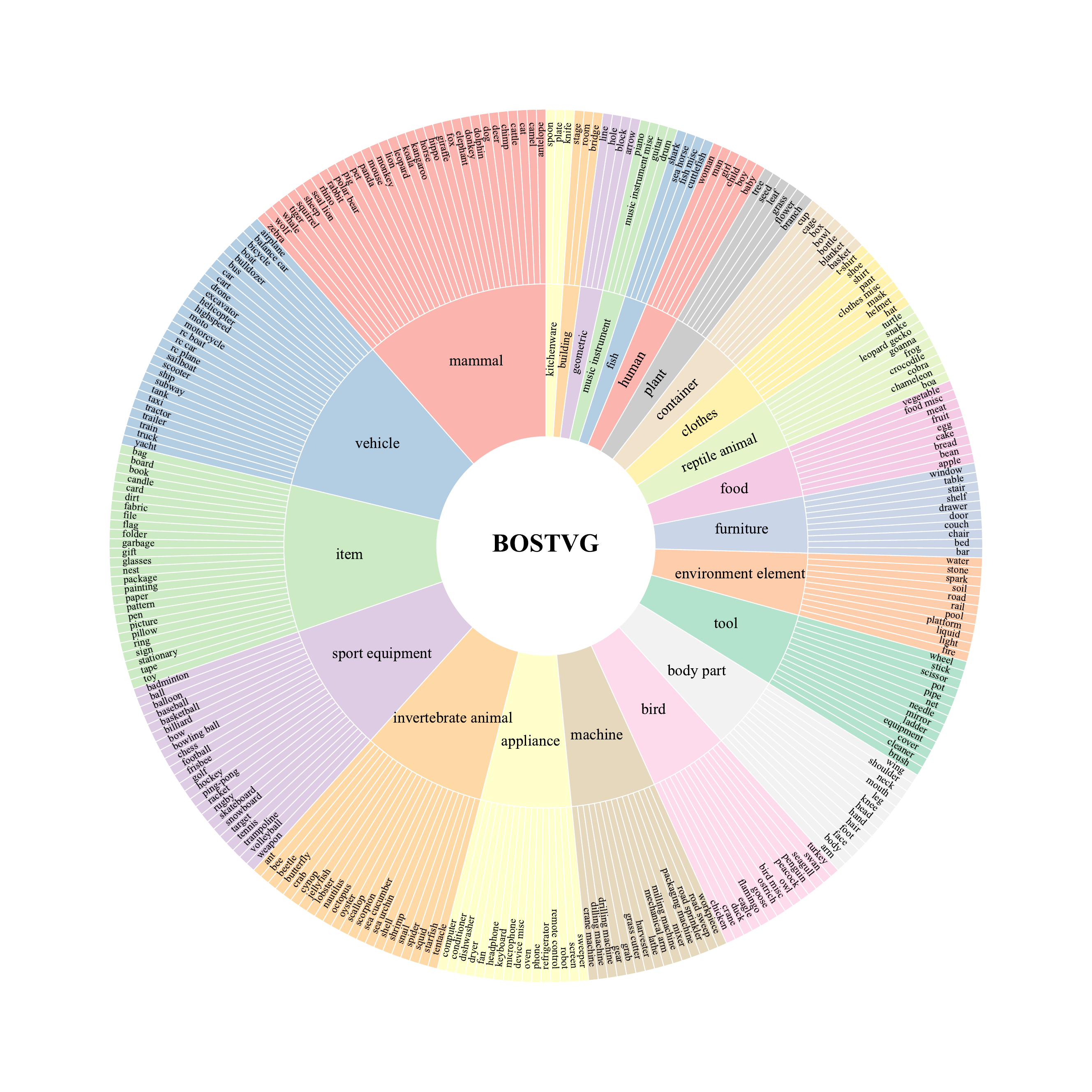}\vspace{-1mm}
    \caption{Category organization of our BOSTVG. The inner circle of the pie chart displays 23 coarser object classes, while the outer circle displays 287 fine object categories. \emph{Best viewed in pdf and by zooming in}.}
    \label{fig:category_pie}\vspace{-3mm}
\end{figure*}

\section{Details of Object Categories}
\label{sec: categories}
BOSTVG contains 287 object categories, aiming to provide a diverse platform for the OmniSTVG task. The categories are organized hierarchically to ensure comprehensive coverage. Specifically, we first collect 23 coarse object classes, comprising ``\emph{Appliance}", ``\emph{Bird}", ``\emph{Body Part}", ``\emph{Building}", ``\emph{Clothes}", ``\emph{Container}", ``\emph{Environment Element}", ``\emph{Fish}", ``\emph{Food}", ``\emph{Furniture}", ``\emph{Geometric}", ``\emph{Human}", ``\emph{Invertebrate Animal}", ``\emph{Item}", ``\emph{Kitchenware}", ``\emph{Machine}", ``\emph{Mammal}", ``\emph{Music Instrument}", ``\emph{Plant}", ``\emph{Reptile Animal}", ``\emph{Sport Equipment}", ``\emph{Tool}", and ``\emph{Vehicle}". Please note, since ``\emph{Human}'' is a special category, we separate it from ``\emph{Mammal}''. After this, we further gather 287 fine categories from coarse classes. Fig. \ref{fig:category_pie} shows the category organization of BOSTVG (\textbf{please zoom in}). We will provide and release the category information with our BOSTVG on our website.

\begin{figure}[thb]
    \centering
    \begin{minipage}{0.475\linewidth}
        \centering
        \includegraphics[width=\linewidth]{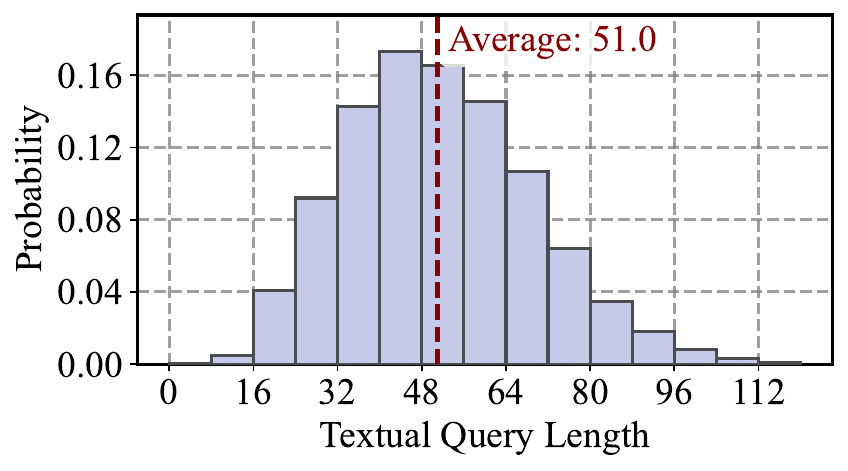}\vspace{-3mm}
        \caption{Distribution of textual query length (in characters)}
        \label{fig:query_len}
    \end{minipage}%
    \begin{minipage}{0.475\linewidth}
        \centering
        \includegraphics[width=\linewidth]{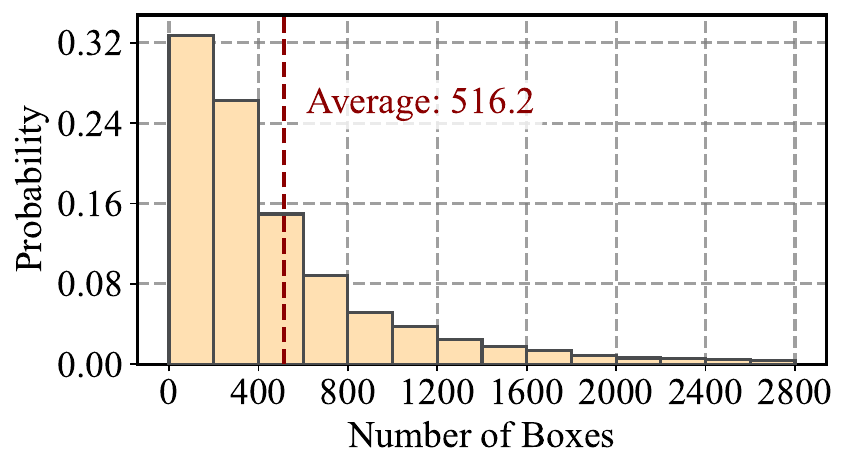}\vspace{-3mm}
        \caption{Distribution of number of boxes in the video}
        \label{fig:box_distribut}
    \end{minipage}\vspace{-5mm}
\end{figure}

\section{More Statistics}

To better understand features of BOSTVG, we further show representative statistics on the textual query length and annotation boxes. Fig.~\ref{fig:query_len} illustrates the distribution of query length, with an average length of 51 characters, which indicates that our dataset offers detailed textual descriptions of target objects. In Fig.~\ref{fig:box_distribut}, we demonstrate the distribution of number of bounding boxes in the sequence, with an average of 516.2 boxes per sequence. From Fig.~\ref{fig:box_distribut}, we can see that our BOSTVG is challenging due to the requirement of localizing more objects.

\begin{figure}[hbt]
    \centering
\includegraphics[width=\linewidth]{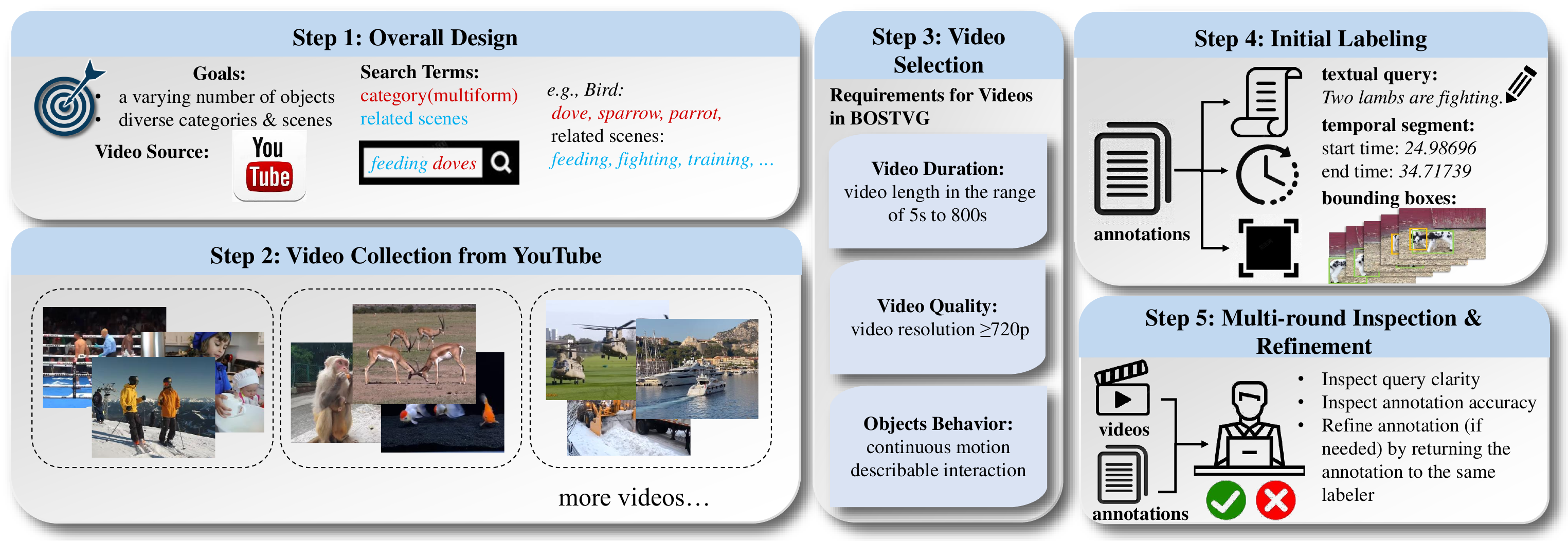}
    \caption{Annotation pipeline for BOSTVG, including five steps, \ie, overall design, video collection, to video selection, initial labeling, and multi-round inspection and refinement.}
    \label{fig:anno_pipleline}
\end{figure}

\section{Construction Pipeline for BOSTVG and Additional Annotation Examples}

The construction of our BOSTVG consists of five steps. In the first step, we determine the overall goal and strategies to search for videos from YouTube. The second step is to collect videos using the strategies in the first step. After this, the third step is to select videos which are qualified for our OmniSTVG task. Following this, the fourth step is to conduct the initial data labeling by our experts. In the final fifth step, we perform multiple rounds of inspection and refinement if needed to ensure high-quality annotations. Fig.~\ref{fig:anno_pipleline} shows the construction pipeline of BOSTVG.

\begin{figure}[htb]
    \centering
\includegraphics[width=0.9\linewidth]{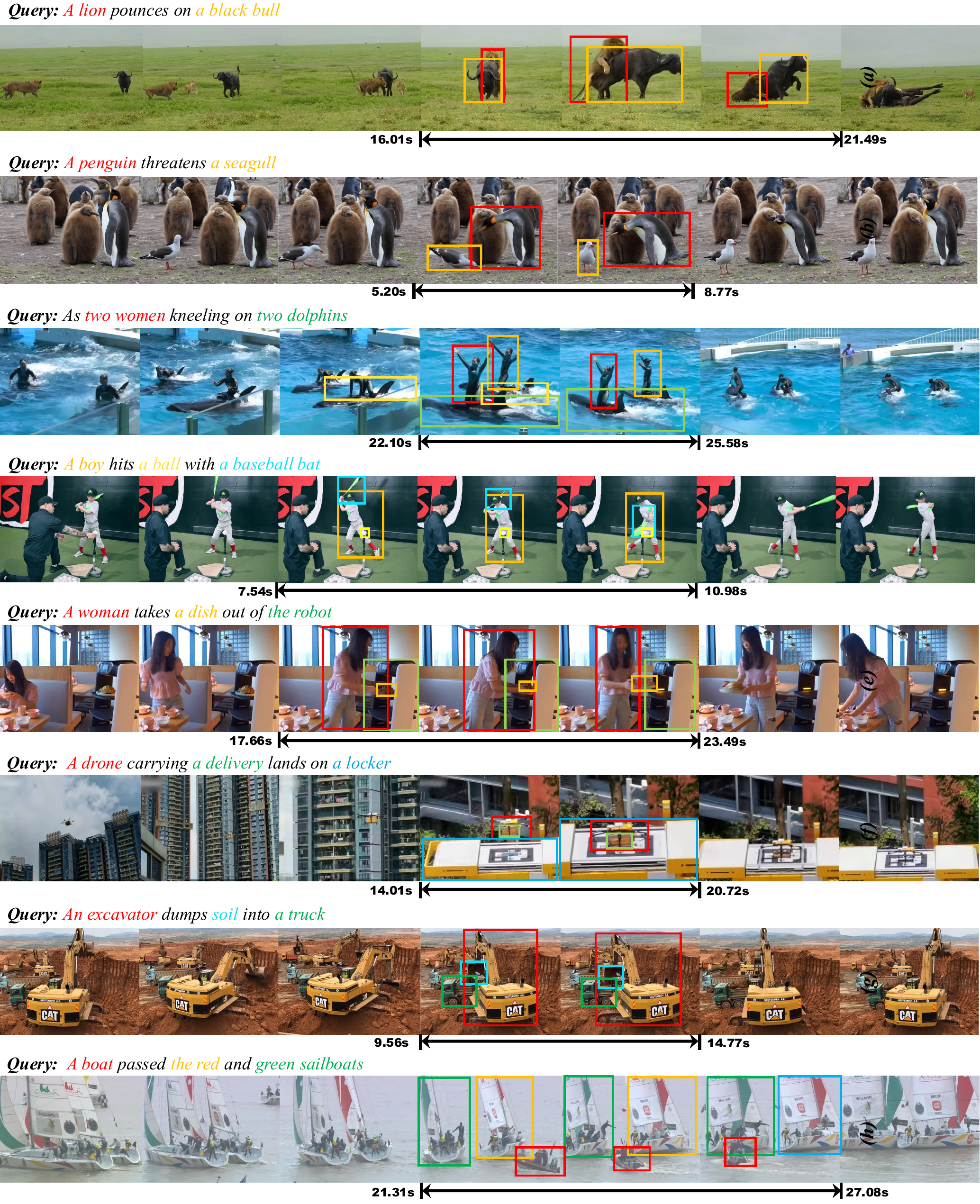}
    \caption{Additional annotation samples on our BOSTVG.}
    \label{fig:anno_samples}
\end{figure}

\section{Detailed Architectures of Modules}

\begin{figure}[!t]
    \centering
\includegraphics[width=\linewidth]{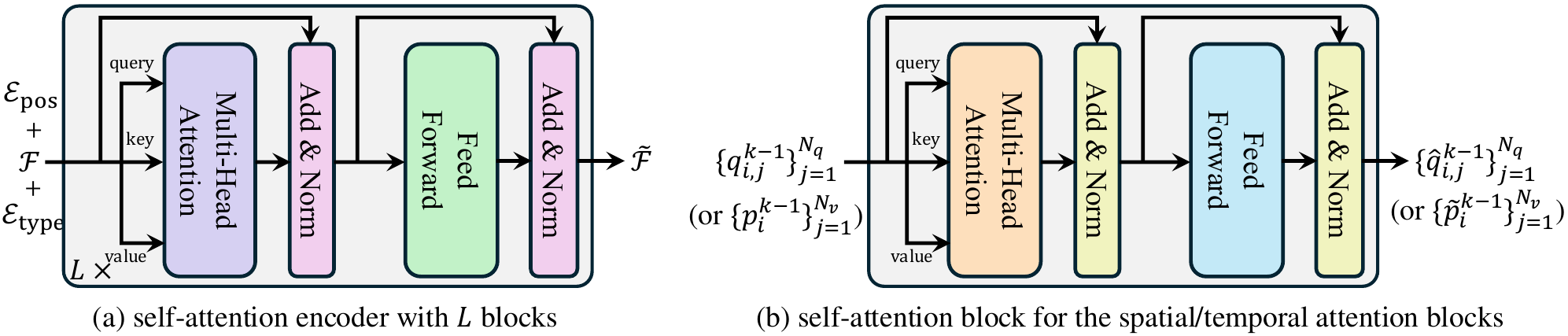}
    \caption{Detailed architectures for the self-attention encoder in (a) and spatial/temporal attention blocks in (b).}
    \label{fig:selfatt}
\end{figure}

In Fig.~\ref{fig:selfatt} (a), we show the architectures for the self-attention encoder, which is composed of $L$ ($L$ = 6) standard self-attention encoder blocks and used to fuse features from multiple modalities. In Fig.~\ref{fig:selfatt} (b)
we show the architectures for the spatial attention block and temporal attention block, which are both implemented with a self-attention block.

\section{Details of Hungarian Loss}

In spatial grounding, we predict the spatial location and class of the bounding box, which can be denoted as 
\begin{equation}\nonumber
\setlength{\abovedisplayskip}{5pt}
\setlength{\belowdisplayskip}{5pt}
\mathcal{Y} = (\mathcal{B}, \mathcal{G}) = \{\{y_{i,j}\}_{i=1}^{N}\}_{j=1}^{N_q}
\end{equation}
where $N$ and $N_q$ denote the number of frames and the number of predicted targets, respectively. $y_{i,j} = (b_{i,j}, g_{i,j})$, where $b_{i,j}$ and $g_{i,j}$ are the spatial location and class of the $j^{\text{th}}$ bounding box in the $i^{\text{th}}$ frame. Suppose that the ground truth for spatial location and class can be denoted as 
\begin{equation}\nonumber
\setlength{\abovedisplayskip}{5pt}
\setlength{\belowdisplayskip}{5pt}
\mathcal{Y}^{*} = (\mathcal{B}^{*}, \mathcal{G}^{*}) = \{\{y_{i,j}^{*}\}_{i=1}^{N}\}_{j=1}^{N_q^{*}}
\end{equation}
where $N_q^{*}$ ($\leq N_q$) denotes the number of ground truth targets. Following DETR~\cite{carion2020end}, to compute the pair-wise matching cost between the prediction results and ground truth, we first utilize the Hungarian matching algorithm to establish a one-to-one correspondence between the prediction and ground truth, as follows,
\begin{equation}\nonumber
\setlength{\abovedisplayskip}{5pt}
\setlength{\belowdisplayskip}{5pt}
\begin{split}
    \mathcal{\hat{Y}} =\mathtt{HungarianMatcher}(\mathcal{Y}^{*},\mathcal{Y})
\end{split}
\end{equation}
where $\mathcal{\hat{Y}}$ represents the successfully matched bounding boxes, while the unmatched bounding boxes are excluded from the loss calculation. After this, we can calculate the loss $\mathcal{L}_h$ as follows,
\begin{equation}\nonumber
\setlength{\abovedisplayskip}{5pt}
\setlength{\belowdisplayskip}{5pt}
\mathcal{L}_h = \lambda_{u}\mathcal{L}_{u}(\mathcal{B}^{*}, \mathcal{\hat{B}}^{*}) + \lambda_{l}\mathcal{L}_{l}(\mathcal{B}^{*}, \mathcal{\hat{B}}^{*}) + \lambda_{c}\mathcal{L}_{c}(\mathcal{G}^{*}, \mathcal{\hat{G}}^{*}) 
\end{equation}
where $\mathcal{L}_{u}$, $\mathcal{L}_{l}$ and $\mathcal{L}_{c}$ are IoU, smooth L1, and binary cross-entropy losses. The parameters $\lambda_{u}$, $\lambda_{l}$, $\lambda_{c}$ are set to $3$, $5$, $1$.

\section{Details of Evaluation Metric}

Following current STVG benchmarks~\cite{ChenMLW19,zhang2020does}, we utilize multiple metrics, including \emph{m\_tIoU}, \emph{m\_vIoU}, and \emph{vIoU@R} for evaluation.
Specifically, \emph{m\_tIoU} aims to assess the temporal localization performance and is calculated by averaging the temporal IoU scores \emph{tIoU} on all test videos. The \emph{tIoU} is calculated as $\frac{|\mathcal{P}_i|}{|\mathcal{P}_u|}$, where $\mathcal{P}_i$ and $\mathcal{P}_u$ represent the intersection and union between the groundtruth and predicted segments, respectively.
\emph{m\_vIoU} is utilized to measure the spatial localization performance, which is calculated by averaging spatial IoU scores \emph{vIoU}. The \emph{vIoU} is calculated as follows,
\begin{equation}\nonumber
\setlength{\abovedisplayskip}{5pt}
\setlength{\belowdisplayskip}{5pt}
    vIoU = \frac{1}{|\mathcal{P}_u|} \sum_{t \in \mathcal{P}_i} (\frac{1}{N_q} \sum_{i \in N_q} \mathtt{IoU}(b^*_{t,i}, b_{t,i}))
\end{equation}
where $b^*_{t,i}$ and $b_{t,i}$ are the groundtruth bounding box and the predicted bounding box of the $i^{\text{th}}$ target object in $t^{\text{th}}$ frame. The \emph{vIoU@R} is defined as the ratio of samples with spatial IoU scores above the threshold $R$.

\section{Qualitative Results}

\begin{figure}[htb]
    \centering
\includegraphics[width=1\linewidth]{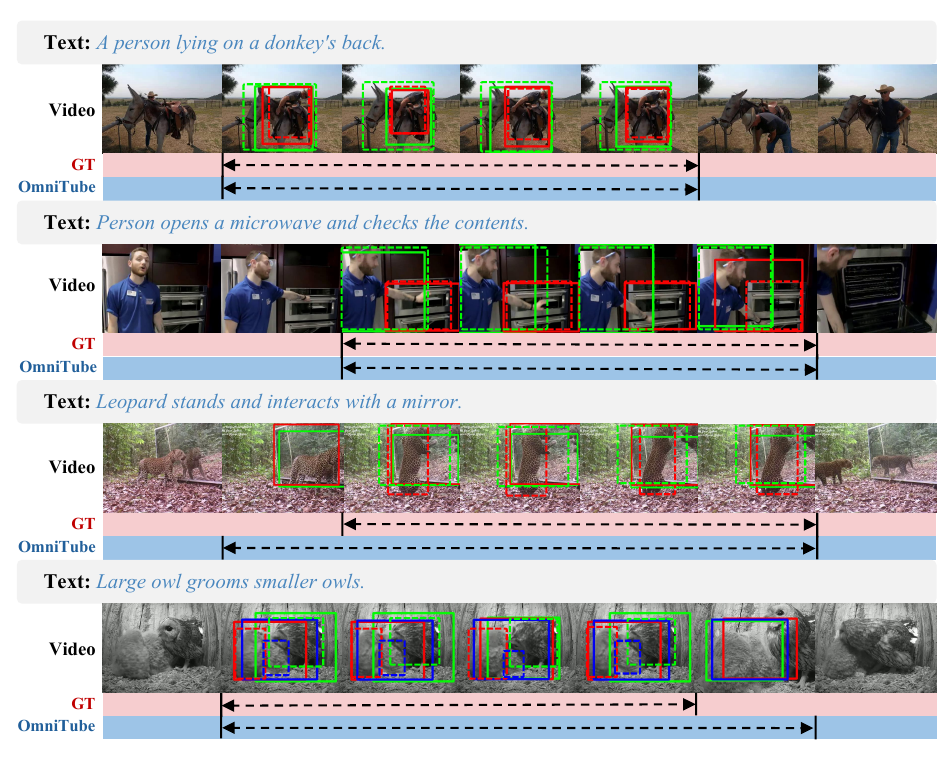}
    \caption{Qualitative results of our method. Our prediction results are visualized with \emph{\textbf{solid-line}} bounding boxes, and the groundtruth boxes are shown in the \textbf{\emph{same color}} with the prediction results but with \emph{\textbf{dashed-line}} bounding boxes.}
    \label{fig:qualitate}
\end{figure}

To further qualitatively validate the effectiveness of our OmniTube, we provide the grounding results of our method in Fig.~\ref{fig:qualitate}. As shown in Fig.~\ref{fig:qualitate}, we can see that, our method can robustly localize all objects mentioned in the textual query, showing its effectiveness.

\end{document}